\documentclass{article}

\usepackage{PRIMEarxiv}

\usepackage[utf8]{inputenc} % allow utf-8 input
\usepackage[T1]{fontenc}    % use 8-bit T1 fonts
\usepackage{hyperref}       % hyperlinks
\usepackage{url}            % simple URL typesetting
\usepackage{booktabs}       % professional-quality tables
\usepackage{amsfonts}       % blackboard math symbols
\usepackage{nicefrac}       % compact symbols for 1/2, etc.
\usepackage{microtype}      % microtypography
\usepackage{lipsum}
\usepackage{fancyhdr}       % header
\usepackage{graphicx}       % graphics
\graphicspath{{media/}}     % organize your images and other figures under media/ folder
\usepackage{caption}
\usepackage{subcaption}

\title{Monocular Cyclist Localization with Convolutional Neural Networks
}

\author{
  Charles Tang \\
  \textit{Primary Author} \\
  Massachusetts Academy of Math and Science \\
  Worcester, MA\\
  \texttt{ctang5@wpi.edu} \\
}

\begin{document}
\maketitle

\begin{abstract}
Cycling is an increasingly popular method of transportation for sustainability and health benefits. However, cyclists face growing risks, especially when encountering large vehicles on the road. This study aims to reduce the number of vehicle-cyclist collisions, which are often caused by poor driver attention to blind spots. To achieve this, we designed a state-of-the-art real-time monocular cyclist detection that can detect cyclists with object detection convolutional neural networks, such as EfficientDet Lite and SSD MobileNetV2. First, our proposed cyclist detection models achieve greater than 90\% mAP (IoU: 0.5), fine-tuned on a newly proposed cyclist image dataset comprising over 20,000 images. Next, the models were deployed onto a Google Coral Dev Board mini-computer with a camera module and analyzed for speed, reaching inference times as low as 15 milliseconds. Lastly, the end-to-end cyclist detection device was tested in real-time to model traffic scenarios and analyzed further for performance and feasibility. We concluded that this cyclist detection device can accurately and quickly detect cyclists and has the potential to improve cyclist safety significantly. Future studies could determine the feasibility of the proposed device in the vehicle industry and improvements to cyclist safety over time.
\end{abstract}

% keywords can be removed

\twocolumn

\section{Introduction}
Cyclists face numerous risks when traveling along urban roads and intersections, and face the most significant risks when encountering semi-trailer trucks because they are large vehicles that can cause severe accidents when drivers have trouble identifying cyclists in their blind spots when making right-hand turns \cite{richter_turning_2017}. 

\begin{figure}[b]
  \centering
  \includegraphics[scale=0.5]{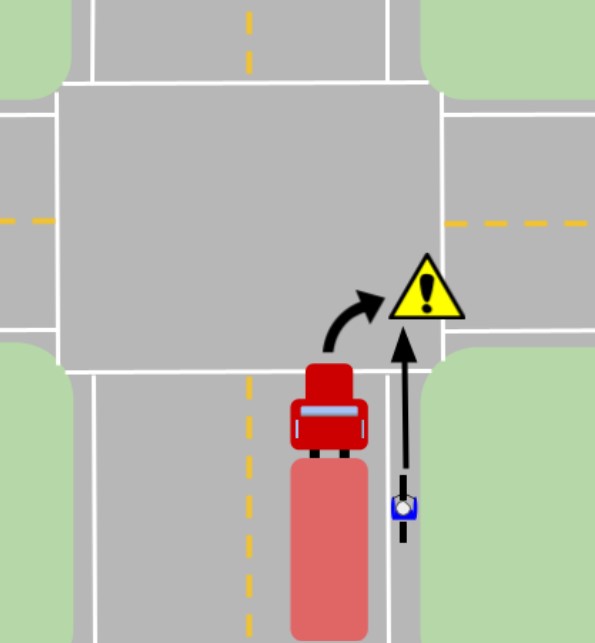}
  \caption{Semi-trailer Truck Right-Hook Turn Scenario.}
  \label{fig:fig1}
\end{figure}

\begin{figure}[b]
\centering
\includegraphics[scale=0.5]{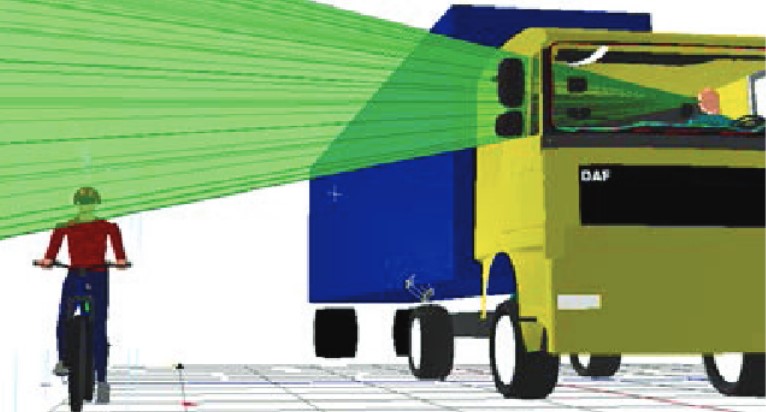}
\caption{Limited Visibility of Cyclists From a Semi-trailer Truck Driver.}
\label{fig:fig2}
\end{figure}

\subsection{Cyclist Safety and Collisions}
Bicycling is a hobby and sport that many enjoy for its fitness benefits and improvement in mental health. Unfortunately, cyclists often face risks when traveling along a road or path with other motorized vehicles; in the U.S., over 1,000 cyclists die in road accidents, and 130,000 more are injured yearly \cite{noauthor_bicycle_2022}. 

Some risk factors that increase the rate of cyclist collisions include places with fast-moving vehicles, intersections, larger vehicles, poor infrastructure, steep terrain, and poor lighting \cite{carvajal_bicycle_2020}. Furthermore, over 71\% of cyclist incidents happen in urban areas, with 30\% of collisions happening at intersections. These cyclist accidents can be attributed to a lack of visual attention, including distraction, which is a leading cause of over 55\% of vehicle-cyclist collisions \cite{jannat_right-hook_2020}. 

One type of collision that frequently occurs between cyclists and vehicles is in right-hook turns (Figures 1 \& 2). Right-hook collisions occur at intersections when right-turning vehicles collide with a cyclist moving straight forwards along the right side of the road. These types of collisions occur when drivers do not detect cyclists in their oncoming blind spots or when cyclists do not observe that a driver is going to take a right turn. Some factors that increase the risk of right-hook crashes include the presence of left-turning oncoming traffic, high cyclist speed and position in a blind spot, and the presence of crossing pedestrians \cite{jannat_right-hook_2020}. 

\subsection{Blind Spots of Semi-trailer Trucks}
Semi-trailer trucks have large blind spot zones on the vehicle's left, right, front, and rear. The right-hand blind spot poses the most danger to vulnerable road users (VRUs) – cyclists, pedestrians, moped riders, and motorcyclists—especially when making right turns \cite{wang_exploring_2022}. Blind spots are a leading cause of truck-cyclist incidents—they account for 45\% of all collisions between bicycles and trucks \cite{wang_exploring_2022}. Additionally, semi-trailer trucks are disproportionately involved in more severe cyclist collisions compared to other vehicles \cite{richter_turning_2017}. When a semi-trailer truck makes a right turn, the blind spot from the right rearview mirror increases as the view is blocked by the trailer's body. This makes turns especially dangerous because VRUs in this blind spot zone cannot be detected and face a high risk of collision and fatality \cite{wang_exploring_2022} (Figure 2).

In the European Union, as a safety measure, semi-trailer truck drivers are instructed to check their blind spots before and during a right turn to check for VRUs. However, a study investigating the glance behavior of truck drivers during right-hand turns shows that drivers only correctly check their blind spot mirrors about 50\% of the time, which could lead to a higher risk of collision between trucks and VRUs \cite{jansen_caught_2022}. 

The recent growth in interest in cycling places a more significant need on safety technologies and addressing concerns to make cycling more inclusive for all \cite{dill_revisiting_2016}. Furthermore, the general opinion on bicycling in urban areas shows that most people are concerned about the safety of cycling. A survey covering the top 50 metropolitan areas found that 51\% of people are interested in cycling but concerned about the risks without improved safety systems \cite{dill_revisiting_2016}. This shows a significant need for widespread safety systems to make cycling more appealing to all.

\subsection{Object Detection}
Two lightweight object detection model architectures that were used in this study are the EfficientDet Lite and the MobileNetV2. The majority of object detection architectures have a feature extractor and a meta-architecture. First, images pass through feature extractors taken from models such as MobileNet, EfficientNet, and ResNet that extract high-level features. Then, the features are passed through a meta-architecture such as an EfficientDet or a Single-Shot Detector (SSD) model \cite{sandler_mobilenetv2_2018} \cite{tan_efficientdet_2020}. These models consist of numerous layers to predict bounding boxes and object classifications from features \cite{liu_ssd_2016}. 

Object detection models are often tested on a set of metrics (proposed by COCO Object Detection) that measure a weighted average of the precisions (the percentage of detections that are correct out of all detections made) and recall (the percentage of detections that were made out of all ground-truth detections) of the model. The cyclist detection model predicts each detection as a positive bounding box prediction for a cyclist at a determined confidence threshold. This metric is called Mean Average Precision (mAP), and it is calculated as a weighted mean over various Intersection Over Union (IoU) overlap thresholds \cite{lin_microsoft_2015}. SSD and EfficientDet models are smaller and generally have faster computational speed but lower mAP on smaller objects than larger models \cite{huang_speedaccuracy_2017}. 

Object detection models are often trained using transfer learning where pre-trained models from large datasets are fine-tuned to specialize in a specific task. One such dataset composed of over 200,000 labeled images and 80 object categories that was used in this study for transfer learning is the Microsoft COCO Dataset \cite{lin_microsoft_2015}. 

\subsection{Blind Spot Systems}
Various cyclist detection or blind spot assistance systems for drivers and vehicles have been proposed. Current blind spot alert systems found in everyday vehicles use ultrasonic sensors. However, these sensors prove too inaccurate for accurately localizing cyclists \cite{zhang_design_2019} \cite{victor_blind_2013}. Summerskill and Marshall (2015) thus proposed a design involving a modification to the semi-trailer truck cab by reducing the overall height and adding window apertures to improve visibility. However, this design lacks portability and raises questions about the costs of implementing such a design \cite{summerskill_development_2015}. Clegg (2012) also attempted to solve the blind spot issue by involving a mechanism that changes the angle of the rearview mirrors when making turns or lane changes \cite{clegg_apparatus_2012}. A drawback of these mirror systems is that they do not provide an active warning system that can alert drivers of cyclists in their blind spots, and drivers may ignore their blind spots when making maneuvers. One piece of literature further discusses a deep-learning solution to the cyclist detection problem involving LiDAR scans and a convolutional neural network, which reached 80\% mAP \cite{saleh_cyclist_2017}. While LiDAR systems for cyclist detection prove to be an accurate solution, implementing LiDAR systems on modern vehicles is very costly and challenging, with prices ranging from thousands to tens of thousands of dollars.

\subsection{Engineering Statement}
Semi-trailer truck drivers often have trouble identifying cyclists in their blind spots when making right-hand turns which can cause cyclist-truck collisions. The overall aim of this project was to engineer a device that can detect cyclists in a truck’s right-rear blind spot and provide alerts for semi-trailer truck drivers. 

\subsection{Engineering Objectives}
The primary concern of this project was to design technology to prevent collisions and injuries that occur when drivers cannot see cyclists. In order to do this, we set three major objectives. (1) Develop a system that can actively detect cyclists with greater than 80\% mAP in a semi-trailer truck’s blind spot. (2) Create warnings for cyclists in a truck’s right-rear blind spot within a two-second interval to reduce the risk of truck-cyclist collision. The speed of the blind spot system is critical in preventing collisions so that drivers can react quickly and appropriately to potential collisions. (3) Make it portable and installable on most semi-trailer trucks. Current blind spot safety systems are not easily deployable and have high-cost barriers that prevent widespread usage. 

This study’s proposed visual cyclist detection device is able to mitigate these pitfalls by employing deep learning computer vision techniques using various object detection architectures trained on a proposed dataset of cyclist images. This design provides accurate, low-cost cyclist detection capabilities and can easily integrate into current technologies. The final device consists of a camera for a continuous video feed input and an onboard mini-computer for real-time detection. 

\section{Methodology}
Cyclist safety is crucial in maintaining safe urban environments. This study developed a system to prevent cyclist collisions and potential fatalities or injuries. Additionally, this study contributes a unique implementation of real-time object detection and testing strategies related to driver assistance systems.

\begin{figure}[b]
\centering
\includegraphics[scale=0.4]{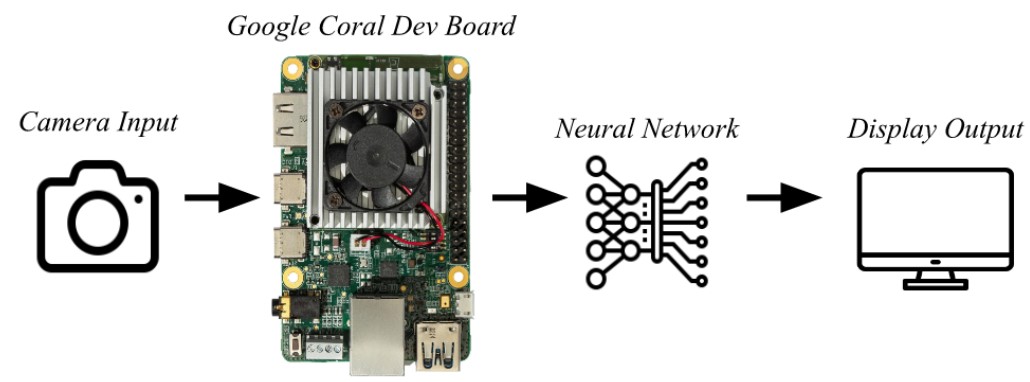}
\caption{Overview of Cyclist Detection Device for Semi-trailer Truck Blind Spots.}
\label{fig:fig3}
\end{figure}

\subsection{Equipment and Materials}
As per this study, the following equipment and software were used to develop the cyclist detection apparatus: an Acer A515 Laptop with 8GB DDR4 RAM and an 11th Gen Intel i5-1135G7 processor, a Google Coral Dev Board, a Logitech C270 USB webcam, an HMTECH 7 inch Raspberry Pi portable monitor, power and data cables, a tripod mount, CAD software, Tensorflow, Google Colab, Tensorflow Object Detection API, Parallel Domain API, Roboflow API, and a bicycle. The hardware materials used in the development of the device are shown in Figure 3. 

\subsection{Dataset Development}
The first step in designing a cyclist detection system using deep learning techniques is to determine or create a dataset with bounding box annotations for cyclists. One such dataset that proves adequate for this task is the CIMAT cyclist orientations dataset \cite{garcia-venegas_safety_2021}. However, current cyclist datasets like CIMAT need more diversity of the bounding box locations of the cyclists as well as edge case scenarios, such as rain, nighttime, or snowy environments. To address this concern, a new dataset was created by combining the CIMAT cyclist dataset with images available from the web and synthetic cyclist images in virtual urban environments from Parallel Domain’s Open Synthetic Dataset for Improving Cyclist Detection \cite{thomas_open_2021}. Then, the collected images were annotated with bounding boxes using Roboflow API and exported as a single-class cyclist Pascal VOC dataset with an 80-20\% train-validation split \cite{dwyer_roboflow_2022} \cite{everingham_visual_2012}. 

\begin{table}[b]
 \caption{Prototype Object Detection Models on COCO Benchmark Dataset.}
  \centering
  \begin{tabular}{ll}
    \cmidrule(r){1-2}
    Object Detection Model & mAP (IoU 50:05:95) \\
    \midrule
    SSD MobileNet V2 & 0.202  \\
    EfficientDet Lite 1  & 0.326  \\
    EfficientDet Lite 2 & 0.418  \\
    \bottomrule
  \end{tabular}
  \label{tab:table1}
\end{table}

\subsection{Cyclist Detection Model}
This cyclist detection system uses innovative technologies to achieve state-of-the-art performance. Among the many object detection architectures, the SSD MobileNet V2 and EfficientDet Lite models were chosen due to their lightweight memory and CPU/TPU usage and strong performance with real-time detection \cite{sandler_mobilenetv2_2018} \cite{tan_efficientdet_2020}. Using the Tensorflow Object Detection API and the Tensorflow Lite Model Maker API, the dataset was used to fine-tune an SSD MobileNet and EfficientDet Lite model for a total of 50 epochs. Table 1 displays the mean average precision (mAP) of the chosen object detection models on the COCO dataset.

After training the object detection models for the single-class cyclist detection task, the model was tested on the validation set from the CIMAT dataset and the newly proposed dataset using average precision object detection metrics. 

\subsection{Deployment and Further Testing}
The trained cyclist detection models were then deployed onto the Google Coral Dev Board with a camera module. The Google Coral Dev Board provides state-of-the-art performance concerning object detection inferences in real-time applications \cite{seshadri_evaluation_2022}. In mobile environments, the onboard Edge Tensor Processing Unit (TPU) can make inferences using the SSD MobileNet V2 with up to 400 frames per second \cite{seshadri_evaluation_2022}. Figure 3 illustrates an outline of the final design. Using the Google Coral Dev Board, the frames per second (FPS) of each object detection model’s inferences were also analyzed. 

\subsection{Real-Time Testing}
The final objective of this study was to design a device that can be portable and installable onto most existing semi-trailer truck cabs. A 3D-printed connector piece was also created using CAD that allowed the camera module to be attached to the right rear-view mirror of a truck. We connected the camera module to the Coral Dev Board and LCD display to be accessible from within a truck cab. 

We then constructed a real-time testing environment with a cyclist, a fixed mount for the blind spot device, and the blind spot device. The cyclist was guided down a straight path to simulate a through-going cyclist. The cyclist detection apparatus was then fixed at three locations to simulate a semi-trailer truck at an intersection as shown in Figure 4: in front of the cyclist, above the cyclist, and behind the cyclist. The first camera was placed at a height of 5 feet, the second camera was placed at a height of 13 feet, and the third camera was placed at a height of 13 feet. The device was tested qualitatively for its effectiveness in detecting cyclists from the viewpoint of a truck cab and for inference speed. 

\begin{figure}[t]
\centering
\includegraphics[scale=0.5]{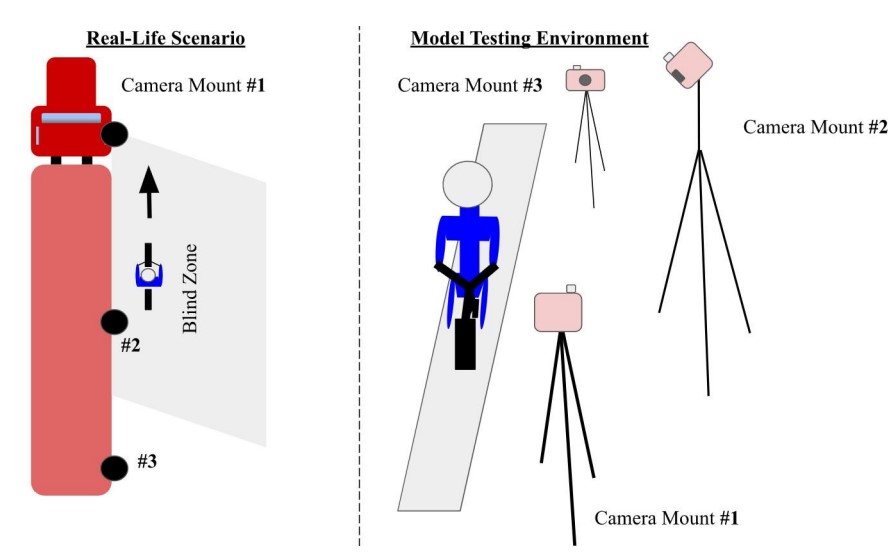}
\caption{Right-Hook Turning Scenario v.s. Model Testing Environment.}
\label{fig:fig4}
\end{figure}

\section{Results}
\subsection{Proposed Cyclist Dataset}
Our proposed auxiliary cyclist dataset contains 20,406 cyclist images with over 40,000 cyclist instances collected from synthetically generated images and collected images from various online sources \cite{thomas_open_2021}. This dataset serves as an extension to the 12,000 cyclist images from the CIMAT cyclist dataset and has a greater focus on edge cases, such as night-time, rainy, and busy urban environments \cite{thomas_open_2021}. Figure 5 shows some example images from the newly proposed dataset with variable lighting and weather conditions. Initially, 6,802 images were collected, and the images were augmented to generate new data points and triple the dataset size. Recommended data augmentation options were applied using the Roboflow API to diversify the dataset and improve model performance: random zoom (0–31\%), hue filter (-25–+25$^{\circ}$), saturation filter (-22\%–+22\%), brightness (-13\%–+13\%), blur (0–0.125px), and noise (0–2\%) (Everingham et al., 2022). Figure 6 shows where cyclist bounding boxes most commonly lie in the dataset images. Figure 7 outlines the number of cyclist instances per image. 

\begin{figure}[b]
\centering
\includegraphics[scale=0.42]{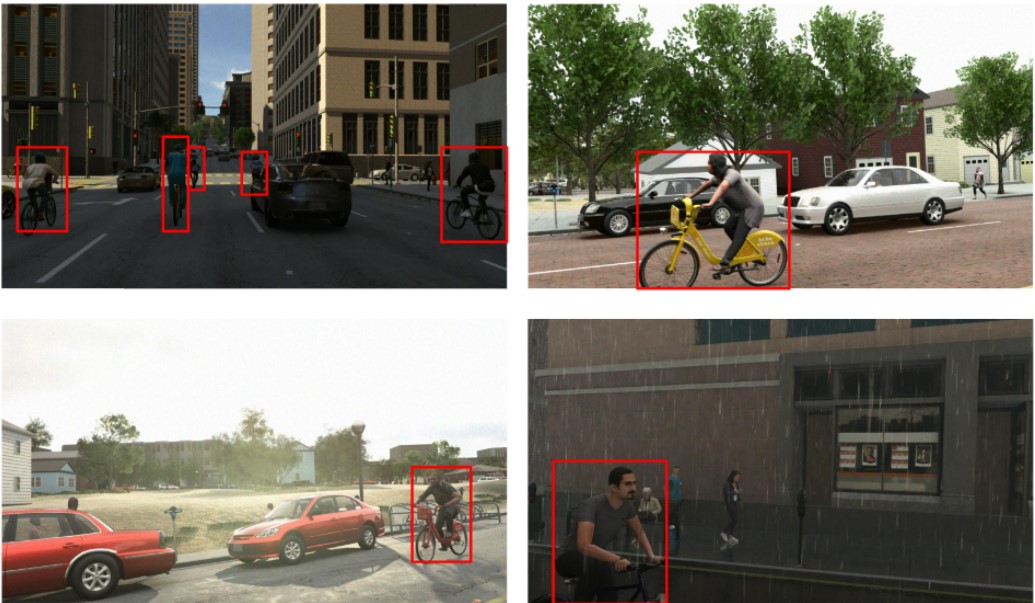}
\caption{Example Images From the Newly Proposed Cyclist Dataset.}
\label{fig:fig5}
\end{figure}

\begin{figure}[t]
\centering
\includegraphics[scale=0.55]{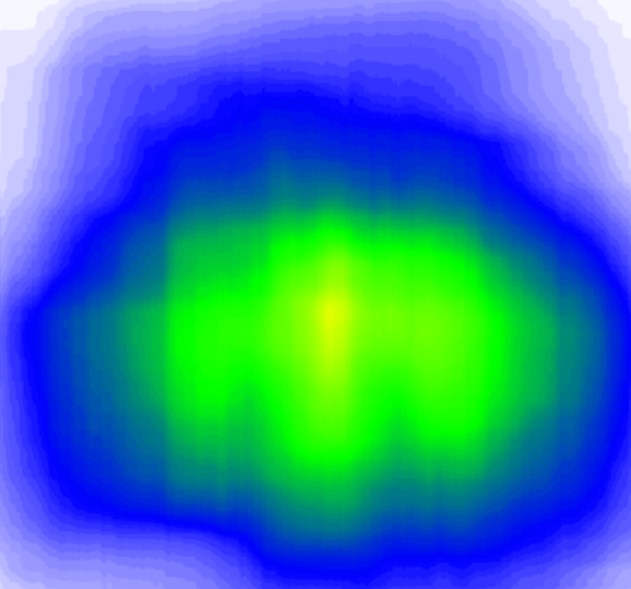}
\caption{Annotation Heatmap of Cyclist Instances in Newly Proposed Dataset.}
\label{fig:fig6}
\end{figure}

\begin{figure}[t]
\centering
\includegraphics[scale=0.55]{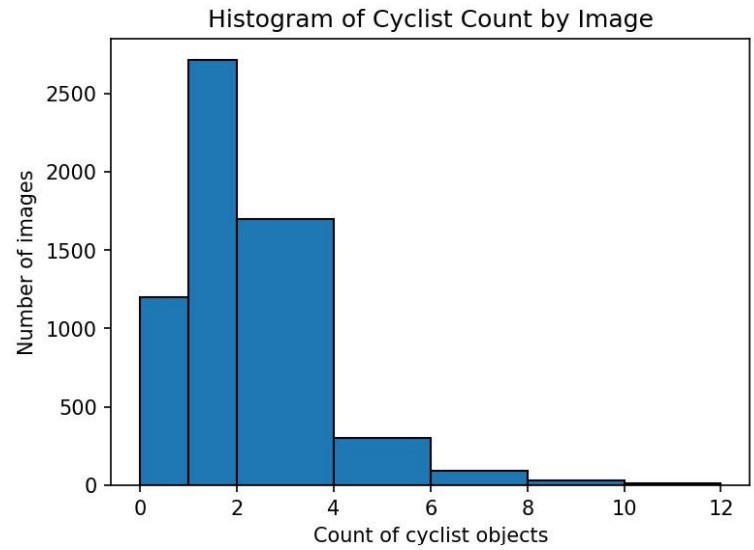}
\caption{Histogram of Cyclist Count Per Image in Newly Proposed Dataset.}
\label{fig:fig7}
\end{figure}

\subsection{Cyclist Detection Models}
The three single-class cyclist detection models were evaluated using average precision (AP) metrics for accuracy. The mAP results are described in Figure 8. The results show that the EfficientDet Lite 1 and the EfficientDet Lite 2 are the highest-performing models, with an mAP of 95.6\% and 95.4\% at the IoU threshold of 0.5, respectively. The SSD MobileNetV2 model performed the weakest at 84.7\% mAP (IoU 0.5). The mAP was measured to determine the effectiveness of this cyclist detection system. Among the various cyclist detection models trained, the three models shown in Figure 8 were trained and tested on both the newly proposed and the CIMAT cyclist dataset and are compatible with the Edge TPU hardware. 

It is assumed that higher-performing AP metrics correlate with more robust performance in real time. Figure 9 illustrates some example predictions made by the cyclist detection models from the validation set. 

\begin{figure}[t]
\begin{subfigure}[t]{0.5\textwidth}
\centering
\includegraphics[width=0.8\linewidth]{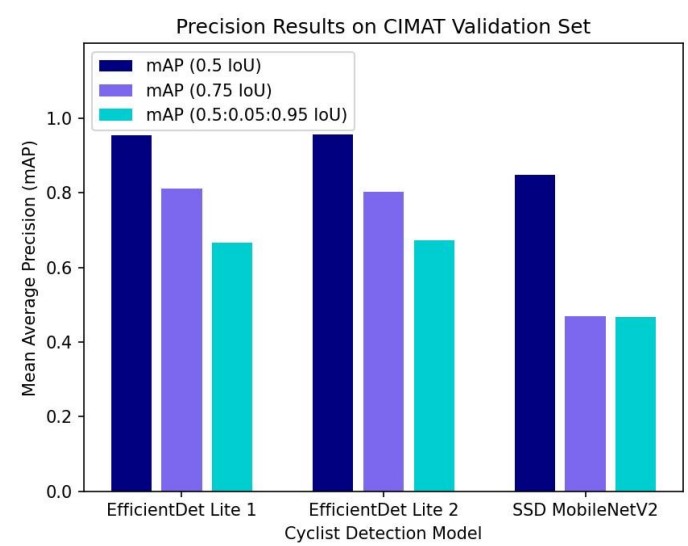}
\subcaption{mAP on the CIMAT Validation Set.}
\end{subfigure} %
\begin{subfigure}[t]{0.5\textwidth}
\centering
\includegraphics[width=0.8\linewidth]{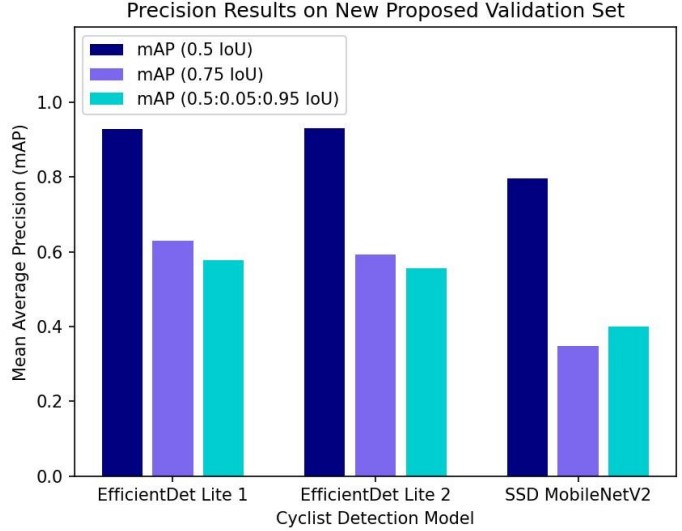}
\subcaption{mAP on the Newly Proposed Validation Set.}
\end{subfigure}
\caption{mAP Results of the Cyclist Detection Models.}
\label{fig:fig8}
\end{figure}

\begin{figure}[b]
\centering
\includegraphics[scale=0.35]{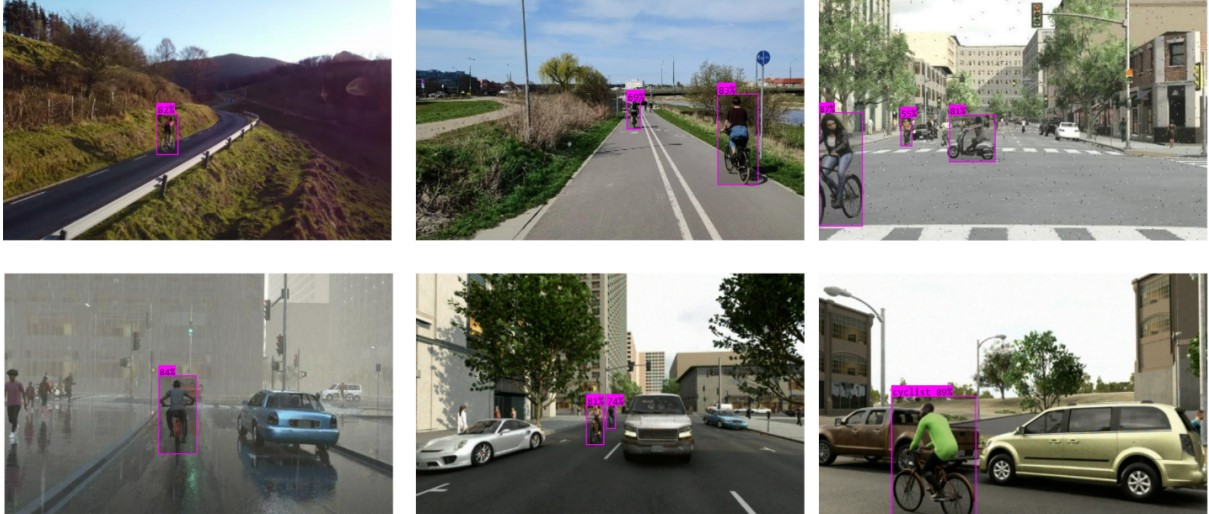}
\caption{Example Cyclist Detections From Newly Proposed Dataset.}
\label{fig:fig9}
\end{figure}

\subsection{Blind Spot Apparatus Design}
A basic CAD model was developed and 3-D printed to enclose the modules of the blind spot alert apparatus and allow it to be portable and installable onto most semi-trailer trucks (Figure 10). The case dimensions are 9” by 5.25” by 5” to enclose a 7-inch monitor, the single board computer, HDMI cables, and power cables. The final design uses a USB power cable that would be plugged into the interior of the truck, as well as a wired webcam that would be attached to the exterior of the truck. The Google Coral Dev Board was also configured to produce audible alerts through an AUX cable when cyclists were detected in the blind zone. 

\begin{figure}
\centering
\includegraphics[scale=0.65]{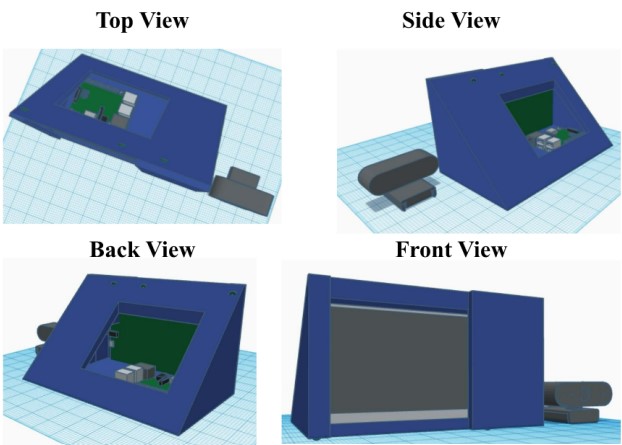}
\caption{CAD Design for Blind Spot Alert Device.}
\label{fig:fig10}
\end{figure}

\subsection{Environment Feasibility Testing}
The testing scenario was designed to model a right-hook turn made by a truck near or in front of a through-going cyclist. The cameras were set up to model a stationary semi-trailer truck with the following dimensions: 80 feet long and 13 feet high. Cameras were fixed along the side of the truck to model the three camera placements illustrated in Figure 4. The SSD MobileNetV2 object detection architecture made the fastest inference times at 70.6 FPS or 0.0142 seconds of inference time (latency). Figure 11 shows the cyclist detection model speeds from real-time testing. Figure 12 and 13 illustrates the testing scenario and example detections of the EfficientDet Lite 1 model.

\begin{figure}
\centering
\includegraphics[scale=0.65]{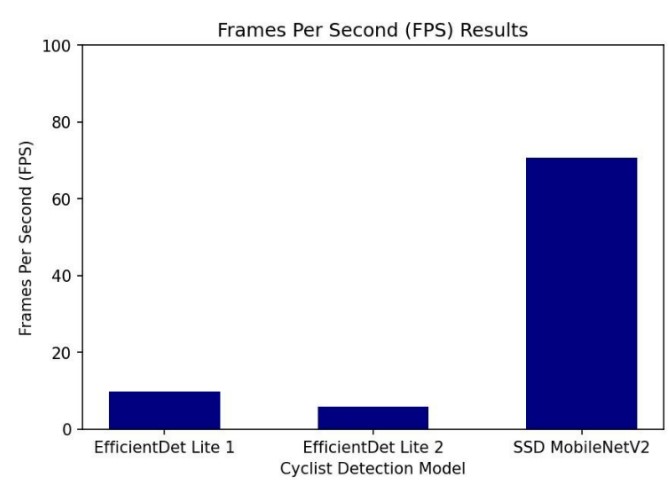}
\caption{Frames Per Second Results From Real-Time Testing.}
\label{fig:fig11}
\end{figure}

\begin{figure}
\centering
\includegraphics[scale=0.4]{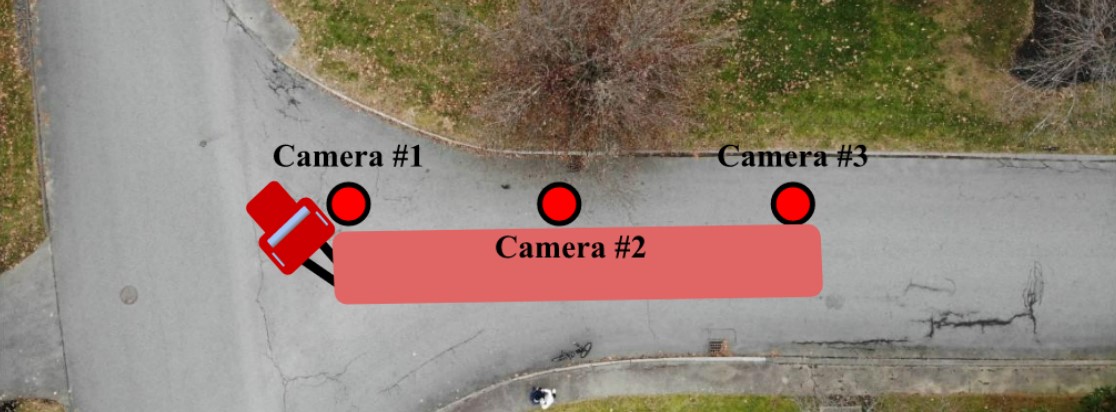}
\caption{Aerial View of Real-Time Testing Environment.}
\label{fig:fig12}
\end{figure}

\begin{figure}
\centering
\includegraphics[scale=0.35]{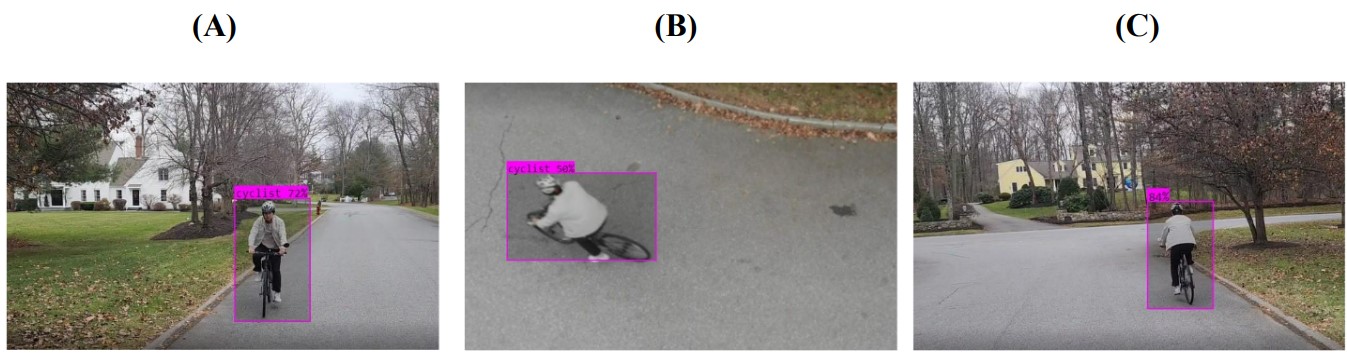}
\caption{Example Cyclist Detections From Real-Time Testing Environment. }
\label{fig:fig13}
\end{figure}

\section{Discussion}
Our study investigated a method of identifying cyclists in the blind spot of a semi-trailer truck and relaying information to a truck driver to prevent truck-cyclist collisions during right-hook turns. This study aimed to develop a portable device that can accurately and quickly detect cyclists within a blind spot. A deep learning approach was used to create a cyclist detection model and was deployed onto a mini-computer for testing. 

First, the cyclist image dataset presented in this study that combines the CIMAT Cyclist dataset and the Open Synthetic Dataset for Cyclist Detection provides more edge-case scenarios and cyclist instances to improve model performance in real-time \cite{garcia-venegas_safety_2021} \cite{thomas_open_2021}. The new auxiliary dataset increases the CIMAT cyclist dataset by 119\%. This dataset aims to train cyclist detection models that can generalize well to various cyclist orientations within various environments. In Figure 6, we can see that most of the annotations are concentrated in the lower central region, with fewer annotations present in the outer regions, as indicated by the contrasting green and blue colors. This suggests that the annotation locations are not evenly distributed across the dataset and that some regions of the dataset may be more heavily annotated than others. Future work to improve this newly proposed dataset is to include images with more diverse locations of cyclists to improve model performance.

From the mAP and inference tests in Figures 8 and 11, the EfficientDet Lite 1 was the most robust, as it maintained a high level of AP while producing inferences at a sufficient speed. The EfficientDet Lite 2 model proved accurate; however, it had a trade-off with an average of 70\% worse performance in time per inference. Despite the much higher inference speeds from the MobileNetV2, the model could not generalize well between the datasets, and the mAP needed to meet the criteria for further use. Furthermore, the two EfficientDet trained models proved to generalize well to various cyclist images from either the CIMAT or Open Synthetic Cyclist dataset with accuracies of 90\% or higher mAP at the IoU threshold of 0.5 (Figure 10). In addition, the trained camera-based cyclist detection models are able to match the LiDAR scan accuracies presented by Saleh et al. (2017) (80\% mAP 0.5:0.05:0.95). These strong performances with the cyclist detection models allow this device to be used in applications beyond right-hand blind spots. Qualitative analysis of the results from the validation set showed that the models sometimes confused pedestrians with cyclists. These confusions could be attributed to the difficult edge cases the models were trained on from the auxiliary dataset. These models also performed adequately at detecting partial cyclists within a frame and cyclists during poor weather conditions, as seen in the example detections from Figure 9, however, the model generally performed better with cyclists in daylight conditions. Furthermore, validation results show no major difference in the effectiveness of detections in urban and rural areas.

Based on the CAD design and the potential camera placements (Figure 4) of the vehicle, it was determined that an attachment of the camera on the right rear-view mirror would be able to (a) perform the strongest in detecting oncoming cyclists (Figure 13: A), and (b) cover a semi-trailer truck’s blind spot most effectively during a right-hook turn at an intersection as outlined by Wang et al. (2022). Furthermore, the system’s active audible and visual alerts address the concerns for driver attention described by Jannat et al. (2020) \cite{jannat_right-hook_2020}. 

In total, the materials for this design cost \$200.00, which would be feasible for the trucking industry to implement as a low-cost, portable, and easily installable system. The installation process was simple—the device needs a USB power outlet, an AUX sound output, and an attachment of the webcam onto the right-rear view mirror and can run cyclist detections almost instantly. Some logistical challenges that may be faced when deploying this device on a wide scale would be making the design components more available to the market. The current shortage in computer components makes it difficult to produce devices such as the one discussed in this study. Another obstacle in making this device more publicly accessible would be making the safety benefits outweigh the investment costs in this technology. 

\subsection{Future Research}
Future work in this field that builds off this study would include real-time testing in various scenarios and widespread testing in the trucking industry. Obtaining feedback from trucking industry users would be valuable in determining the feasibility of the proposed device. Furthermore, additional technologies such as LiDAR and ultrasonic detectors may be developed in conjunction with the proposed device to improve robustness. These avenues of work could be explored to reduce truck-cyclist collisions further. Furthermore, object tracking techniques and detailed risk assessment techniques could be used as discussed by Garcia-Venegas et al. (2021) \cite{garcia-venegas_safety_2021}. Additional techniques using multi-perspective cyclist detection or diffusion object detection models should be investigated to compare to the object detection models proposed in this work \cite{chen_diffusiondet_2022}. Future work could also employ vision transformers to improve awareness of cyclists in a vehicle’s blind spot and improve cyclist detection accuracy \cite{dosovitskiy_image_2021}. 

\subsection{Limitations}
Limitations of this study include the absence of grant funding, computational resources to train deep learning models, and software capabilities to model risk scenarios. It would be most advantageous to test this device on a semi-trailer truck or a similarly-sized vehicle; however, this study could not obtain one for testing purposes.

\section{Conclusion}
The objective of this study was to design an accurate visual-based blind spot warning system for semi-trailer truck drivers to reduce the number of truck-cyclist collisions at right-hook turns. To achieve this, the study set three specific objectives: (1) to develop a system that can actively detect cyclists with high accuracy in a semi-trailer truck's blind spot; (2) to create warnings for cyclists in a truck's right-rear blind spot within a short time interval to reduce the risk of collision; and (3) to design a system that is portable and easily installable on most semi-trailer trucks.

In conclusion, the development of an accurate visual-based blind spot warning system for semi-trailer truck drivers can significantly reduce the number of truck-cyclist collisions. By using state-of-the-art lightweight deep learning architectures and a newly proposed cyclist image dataset, the object detection model was able to locate and detect cyclists with high mAP. The device was tested in real-time and demonstrated good performance and feasibility in a model traffic scenario. Further work is needed to optimize and refine the device, but this study shows promise in improving the safety of VRUs such as cyclists.

\section*{Declarations}
\subsection*{Data Availability}
The code to train the cyclist detection models, the models, and the CAD model can be found at \url{https://github.com/charlestang06/BikeDetector}. The auxiliary dataset can be found at \url{https://app.roboflow.com/bicycle-detection/bike-detect-ct}.

\subsection*{Acknowledgments}
A special thanks to Mr. Medeiros and Dr. Crowthers, who both advised my project, and Grant Perkins, who helped set up some of the software used. Some additional guidance was received from Spencer Chang and Prof. Xinming Huang at WPI. 

\subsection*{Competing Interests}
This study declares no competing interests. This research received no external financial or non-financial support. There are no additional relationships or activities to disclose. 

%Bibliography
\bibliographystyle{unsrt}  
\bibliography{references}

\end{document}